%% file: exCPRArxiv.tex
\documentclass{article}
\input{math_commands.tex}

\usepackage[nonatbib,preprint]{neurips_2023}

\usepackage{hyperref}       

\usepackage[utf8]{inputenc} 
\usepackage[T1]{fontenc}    

\usepackage{url}            
\usepackage{booktabs}       
\usepackage{amsfonts}       
\usepackage{nicefrac}       
\usepackage{microtype}      
\usepackage{graphicx}
\usepackage{amsmath}
\usepackage{float}
\usepackage[english]{babel}
\usepackage{booktabs}
\usepackage[table]{xcolor}
\usepackage{dsfont}
\usepackage{caption}
\usepackage{subcaption}
\usepackage{paralist}

\usepackage{amssymb}
\usepackage{mathtools}
\usepackage{amsthm}
\usepackage{algorithm2e}[ruled]
\RestyleAlgo{ruled}

\newtheorem{theorem}{Theorem}[section]

\newtheorem{lemma}[theorem]{Lemma}
\newtheorem{corollary}[theorem]{Corollary}
\newtheorem{definition}[theorem]{Definition}

\newcommand\norm[1]{\left\lVert#1\right\rVert}
\newcommand{\defeq}{\stackrel{\text{def}}{=}}
\newcommand{\B}{\bfseries}
\definecolor{lightgray}{gray}{0.9}

\title{Minimizing Chebyshev Prototype Risk Magically Mitigates the Perils of Overfitting}

\author{%
  Nathaniel Dean\\
  Department of Computer Science\\
  University of Miami\\
  Coral Gables, FL \\
  \texttt{nxd551@miami.edu} \\
  \And
  Dilip Sarkar \\
  Department of Computer Science \\
  University of Miami \\
  Coral Gables, FL \\
  \texttt{sarkar@cs.miami.edu} \\
}

\begin{document}

\maketitle

\begin{abstract}
Overparameterized deep neural networks (DNNs), if not sufficiently regularized, are susceptible to overfitting their training examples and not generalizing well to test data.  To discourage overfitting, researchers have developed multicomponent loss functions that reduce intra-class feature correlation and maximize inter-class feature distance in one or more layers of the network.  By analyzing the penultimate feature layer activations output by a DNN's feature extraction section prior to the linear classifier, we find that modified forms of the intra-class feature covariance and inter-class prototype separation are key components of a fundamental Chebyshev upper bound on the probability of misclassification, which we designate the Chebyshev Prototype Risk (CPR).  While previous approaches' covariance loss terms scale quadratically with the number of network features, our CPR bound indicates that an approximate covariance loss in log-linear time is sufficient to reduce the bound and is scalable to large architectures.  We implement the terms of the CPR bound into our Explicit CPR (\emph{exCPR}) loss function and observe from empirical results on multiple datasets and network architectures that our training algorithm reduces overfitting and improves upon previous approaches in many settings.  Our code is available  \href{https://github.com/Deano1718/Regularization_exCPR}{here}. 
\end{abstract}


\section{Introduction}\label{Intro}

Deep neural networks (DNNs) have shown to be effective pattern extractors and classifiers, resulting in remarkable and increasing performance in visual classification tasks over the last two decades.  A large portion of the performance increase may be attributed to improved architectures, increased scales, and larger quantity training sets, but these classifiers are still at risk to the phenomenon of overfitting to a particular training set, which equates to rote memorization of specific examples and decreased generalization.

Many strategies have been developed over the years to regularize networks during the training process such as data augmentation, weight decay \cite{hintonWeightDecay}, dropout \cite{hinton2012Dropout}, and batch normalization \cite{ioffeBN}, which have now become standard practices in deep neural network training.  In parallel to these training strategies, several efforts have looked at augmenting the standard cross-entropy loss function with additional terms that seek to decorrelate learned feature representations and eliminate redundant weights.

We analyze the mathematical basis for removing the covariance between feature representations and in doing so, transfer the concept of the \emph{class prototype} from the field of deep metric learning \cite{mensinkMetricLearning} into our derivations.  We utilize the class prototype, which is the class' mean feature vector, to derive Chebyshev probability bounds on the deviation of an example from it's class prototype and to design a new loss function that we empirically show to excel in performance and efficiency compared to previous algorithms.  In this paper, we make the following contributions:

\begin{itemize}
	\item A theoretical framework based on Chebyshev probability bounds under which regularization and related training techniques can be analyzed.  The bound admits a new optimizable metric called Chebyshev Prototype Risk (CPR), which bounds the deviation in similarity between the penultimate features of an example and its class prototype.
	\item A new loss function augmented with CPR terms that reduces intra-class feature covariance while keeping inter-class feature vectors separated.  The loss function properties generalize to unseen examples and reduce the risk of overfitting.
	\item To the best of our knowledge, the first regularization algorithm to effectively optimize feature covariance in log-linear time and linear space, thus allowing our algorithm to scale effectively to large networks. 
\end{itemize}  

\subsection{Preliminary Notations}

There exists a generic classifier $f(\vx ; \theta)$ that maps an input vector $\vx\in \R^M$ to a vector $\vy \in \R^K$.  We select the parameters, $\theta$, of this classifier as an estimator using a learning algorithm $\mathcal{A}(\mathcal{D})$ that takes as input a labeled training set $\mathcal{D} = \{(\vx_1,\vy_1),  (\vx_2,\vy_2) ... (\vx_N,\vy_N)\}$.  This classifier can be broken down into the composition of a \emph{feature extraction} function $g:\R^M \rightarrow \R^J$, which outputs a feature vector $\vv \in \R^J$ in the feature vector layer, and a \emph{feature classification} function $h: \R^J \rightarrow \R^K$.  Therefore, $\vy = f(\vx;\theta) = h(g(\vx; \theta_{g}) ; \theta_{h}) = h(\vv ; \theta_{h})$, where $\theta_{g}$ and $\theta_{h}$ are the disjoint parameter sets on which $g$ and $h$ are dependent, respectively.  Unless otherwise stated, the norm $||\ \cdot\ ||$ will refer to the $L_2$-norm.

There exists for each category a learnable, class representative feature vector $\vp_k\in \R^J$ that is learned during training and minimizes some feature similarity based loss function between itself and all of its constituent examples in class $k$.  Thus, there is a defined set of class prototype feature vectors $\left\{\vp_1, ... ,\vp_K\right\}$.

\begin{figure}[h]
	\centering
	\includegraphics[width=0.6\linewidth]{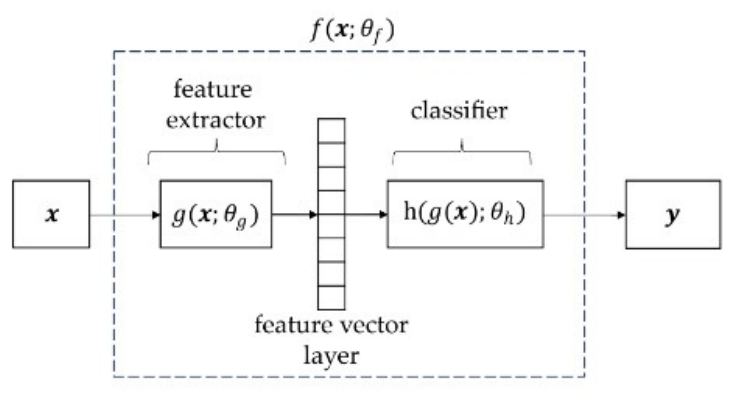}
	\caption[gm]{Neural network split into a feature extractor and classifier (last fully connected layer) acting on an input $\vx$.  Class prototypes are learned feature vectors that comprehensively represent each class' trained features. \\}
	\label{fig:flowchart}
\end{figure}

\section{Related Work}

A rich history exists for regularization algorithms that optimize feature decorrelation or separation in one or more layers of a model such as \cite{squent,arcfaceDeng,ayinde,huang2017orthogonal,pereyra2017regularizing, liu2017largemargin,Wan2013RegularizationON, hinton2012Dropout}.  Almost all of these methods work off the notion that feature decorrelation and separability are beneficial to the test performance of a network.  We show that this test performance and feature separability hypothesis is well-supported by basic probability inequalities, but the precise mathematical form of our CPR bound differs from these previous works.  We now present the details of several previous algorithms in this field.  

\paragraph{DeCov} The loss function in \cite{cogswellDecov} discourages feature correlations at selected layers of a deep neural network by implementing a penalty loss term on intra-batch covariance.  For a given mini-batch, the loss defines the batch feature covariance as,

\begin{align}
	\begin{split} \label{eq:1}
		C_{i,j} =  \frac{1}{N}\sum\limits_{n}^{N}(h_i^n - \mu_i)(h_j^n - \mu_j) 
	\end{split}
\end{align} 

where $N$ is the number of samples in the batch, $h$ are the feature activations at the selected layer, and $\mu$ is the sample mean of feature activations for the batch.  Armed with this definition, they augment the standard cross-entropy loss function with the following regularization term:

\begin{align}
	\begin{split} \label{eq:2app}
		\mathcal{L}_{DeCov} =  \frac{1}{2}(||C||_F^2 - ||diag(C)||^2)
	\end{split}
\end{align} 

where $||\cdot||_F$ is the Frobenius norm.  This term reduces the magnitude of the off-diagonal terms of the observed batch feature covariance matrix regardless of class.  The hyperparameter $w$ is the relative weighting of the DeCov loss component compared to the cross-entropy loss.

\paragraph{OrthoReg} In \cite{rodríguez2017regularizing}, the authors design a weight update technique that directly controls the cosine similarity between weight vectors throughout the neural network.  The regularization cost function introduced is,

\begin{align}
	\begin{split} \label{eq:3app}
		C(\theta) =  \frac{1}{2} \sum\limits_{i=1}^n \sum\limits_{j=1,j \ne i}^n \left( \frac{\langle \theta_i , \theta_j \rangle}{||\theta_i||||\theta_j||} \right)^2
	\end{split}
\end{align} 

where $\theta_i$ is the weight vector connecting to neuron $i$ of the next layer, which has n hidden units.  Two different weight update rules are derived from this cost function: one that regularizes both positive and negatively correlated weights and another that penalizes only positive correlations.  The authors hypothesize that negative correlations should not be penalized as they aide in generalization.

\paragraph{Squentropy}  The squentropy loss \cite{squent} augments the standard cross-entropy loss term to minimize the squared $L_2$-norm of the non-true label logits:

\begin{align}
	\begin{split} \label{eq:4app}
		\mathcal{L}(\vx_{k},y_k) =  -\mathds{1}(j = y_k)\sum\limits_{j}^K log(p_j) + \frac{1}{K-1} \sum\limits_{i=1,i\ne k}^K z_i^2
	\end{split}
\end{align}

where $p_j$ is the $j^{th}$ output of the softmax function $\sigma(\vz)$ acting on the logit vector $\vz \in \R^K$.  Since the cross-entropy loss tends to maximize the logit value of the true class, constricting the non-true logits towards zero magnitude tends to create inter-class angular orthogonality at the logit level, which carries over well to the feature space $g(\vx)$.

\section{Theoretical Framework}

For each class, we define a feature vector called the class prototype.  Succinctly, given the feature extraction function $g(\cdot; \theta_g)$, the prototype $\vp_{k}$ for class $k$ is defined as the $J$ dimensional vector in the feature layer of a DNN whose elements are the arithmetic means of all $N_k$ feature vectors from class $k$ examples in the training set: 

\begin{align}
	\begin{split}
		\vp_{k,j} = \frac{1}{N_k}\sum_{n=1}^{N_k} \ g(\vx_{n,k})_j \ \ \ j=1...J
	\end{split}
\end{align} 

When we compute the cosine similarity between two vectors, we are equivalently first normalizing each vector to unit magnitude and taking their dot product, which returns the cosine of the angle between them.  Since the unit vector is normalized by its magnitude, each individual component is bound to the interval $[0,1]$ for non-negative features and we can then compare vectors of any magnitude on the common basis of their relative angular position.  From this viewpoint, the relative \emph{shapes} of each vector determine their similarity, rather than their magnitudes.

\subsection{Chebyshev Prototype Risk (CPR)} \label{sec:cheb}

Previous studies have shown that deep neural networks classify new examples largely based on their angular position in feature space to the class mean feature vector \cite{Kansizoglou2022DeepGeometry} \cite{Seddik_Tamaazousti_2022_Means}.  We adopt this view of DNN classification behavior and start our theoretical framework by asking, if a class prototype feature vector is an ideal, mean representation for it's class examples and if an example is classified correctly if it is similar in angular feature space to a prototype, then how can we compute or bound the probabilistic risk that a randomly drawn new example is \emph{dissimilar} to its class prototype, thus potentially resulting in a misclassification?

We first define the cosine similarity function for two generic vectors:
\begin{align}
	\begin{split} \label{eq:4_4}
		CS(\vv,\vu) \defeq  \frac{\vv \cdot \vu}{||\vv||||\vu||} 
	\end{split}
\end{align}
We now define a similarity based measure that captures the average \emph{dissimilarity} in feature space of all $K$ prototype feature vectors:

\begin{definition}
	Given a sufficiently trained classifier with low empirical risk, $f(\vx,\theta)$, and a set of $K$ prototype feature vectors of dimension $J$, $\left\{\vp_1, ... ,\vp_K\right\}$, each being a mean representation of a corresponding class $k$, the prototype dissimilarity value for a class $k$, $DS_k \in [0,1]$ is:
	\begin{align}
		\begin{split} \label{eq:2a}
			DS_k &\defeq 1 - \frac{1}{(K-1)} \sum\limits_{i \ne k}^{K-1} CS(\vp_k,\vp_i)\\
		\end{split}
	\end{align}
\end{definition}
The quantity $DS_k$ reflects the average angular spacing of a class prototype from all other class prototypes; maximizing $\sum\limits_{k}^{K} DS_k$ encourages non-negative class prototype feature vectors to be mutually orthogonal after training: if $\forall (k \ne i), CS(\vp_k,\vp_i) \rightarrow 0$, then $DS_k \in [0,1] \rightarrow 1$.  Intuitively, if the angular space between class-mean prototypes can be maximized and generalized to the test set, it would benefit overfitting for the same level of feature variation. 

Next, we are interested in relating the intra-class feature variation to the angular separation between an example's features and its class-mean prototype.  We view the angular distance between a random feature vector and a class prototype as a surrogate for classification probability: the further an example's features separate from a class prototype, the less likely it will be classified into that category.  Lemma \ref{lem:1} uses Chebyshev's inequality to bound the probability that an example's features angularly deviate more than the prototype dissimilarity value from its expected angle to its prototype.  

\begin{lemma} \label{lem:1}
	Given a sufficiently trained classifier with low empirical risk, $f(\vx,\theta)$, a fixed prototype feature vector $\vp_k$ of dimension $J$, which is the mean feature vector of a corresponding class $k$, a prototype dissimilarity value $DS_k$, a feature vector $\vv_k$ for a randomly drawn class $k$ input example, thus resulting in a random vector $X_k$ of the $J$ random variables $v_1p_1...v_Jp_J$, and the corresponding positive semi-definite covariance matrix of $X_k$: $S_k \in \R^{J\times J}$, then the following inequality holds:
	\begin{align}
		\begin{split} \label{eq:3a}
			Pr\big[\big|CS(\vv_k,\vp_k) - &\E\big[CS(\vv_k,\vp_k)\big]\big| \ge DS_k\big] \le \frac{\mathbf{1}^T S_k \mathbf{1}}{DS_k^2}\\
		\end{split}
	\end{align}
	where $\mathbf{1} \in [1]^J$ is the ones vector.
\end{lemma}

The inequality in the above Lemma is the two-tailed Chebyshev's inequality \cite{RossStats}. Intuitively, as the similarity between an example's unit feature vector and a prototype unit vector from class $k$ decreases, the chance of that example being classified into $k$ decreases.  For a given prototype $\vp$, there also exist $(K-1)$ dissimilar prototypes towards which an example could become more similar as it deviates from the original prototype.  See Section \ref{sec:AppLem1} for a full proof of Lemma \ref{lem:1}. 

The above Lemma is for a two-sided bound that bounds the probability in both tails of the distribution.  However, if our random variable deviates in one direction only, which is often the case for intra-class cosine similarity measures, we can derive an improved version of Chebyshev called the Chebyshev-Cantelli inequality, which is stated as follows.

\begin{corollary} \label{cor:1}
	If Lemma \ref{lem:1} holds, then the following inequality holds:
	\begin{align}
		\begin{split} \label{eq:13a}
			Pr\big[CS(\vv_k,\vp_k) - &\E\big[CS(\vv_k,\vp_k)\big] \le -DS_k\big] \le \frac{\mathbf{1}^T S_k \mathbf{1}}{\mathbf{1}^T S_k \mathbf{1} + DS_k^2}\\
		\end{split}
	\end{align} 
	where $\mathbf{1} \in [1]^J$ is the ones vector.
\end{corollary} 

The bound in Corollary \ref{cor:1} is tighter and numerically more stable than the two-sided version.  We gain an extra non-negative term in the denominator, $\mathbf{1}^T S_k \mathbf{1}$, which tightens the bound versus the two-sided version.  Additionally, the maximum value of the right-hand side is 1 for positive $\mathbf{1}^T S_k \mathbf{1}$ regardless of the value of $DS_k$.  This is not the case for the two-sided version when $DS_k \rightarrow 0$ and $\frac{\mathbf{1}^T S_k \mathbf{1}}{DS_k^2} \rightarrow \infty$.  To reduce the bound of either form, however, we must minimize $\mathbf{1}^T S_k \mathbf{1}$ and maximize $DS_k$.  See Section \ref{sec:AppCor1} for a full proof of Corollary \ref{cor:1}. 

Still, a key question is how well these properties \emph{generalize} to the overall population of examples $\mathcal{P}$ and transfer to the \emph{test set}.  As our goal, we wish to design an efficient algorithm that can reduce the right hand sides of Lemma \ref{lem:1} and Corollary \ref{cor:1} for each class in population $\mathcal{P}$, which will reduce the probability that our classifier misclassifies on the test set.  From our probability bounds, we define a \emph{Chebyshev Prototype Risk} (CPR) metric that is in the interest of the classifier to minimize during training and is defined as follows.
\begin{definition} \label{def:1}
	Given a randomly drawn feature vector $\vv_k$ with true label $y=k$ and a fixed class mean prototype feature vector $\vp_k$, let the Chebyshev Prototype Risk (CPR) for $\vv_k$ be defined by:
	\begin{align}
		\begin{split}
			Chebyshev\ Prototype\ &Risk\ (CPR) \defeq \frac{\mathbf{1}^T S_k \mathbf{1}}{DS_k^2} \\
		\end{split}
	\end{align}
	where $S_k$ is the semi-positive definite covariance matrix with elements \\ $S_{k,i,q} = \mathrm{Cov}(\hat{v}_{k,i} \hat{p}_{k,i}, \hat{v}_{k,q} \hat{p}_{k,q})$. 
\end{definition}

\section{Approach and Algorithm}

We contribute a training algorithm that explicitly and efficiently minimizes CPR.  To accomplish this task, we minimize a multi-component loss function composed of the cross-entropy loss, computed at the output layer of the classifier, and of several CPR relevant loss components computed \emph{at the feature vector layer}.  For each drawn mini-batch of examples during training, our computed loss function is:

\begin{align} \label{eq:5}
	\begin{split}
	\textrm{\textbf{Feature Vector Layer Loss: }} &\mathcal{L}_{\vv} = \beta \mathcal{L}_{proto}  + \gamma \mathcal{L}_{cov} + \zeta \mathcal{L}_{CS} \\
	\textrm{\textbf{Output Layer Loss: }}&\mathcal{L}_{\vy} = \mathcal{L}_{CE} \\
	\textrm{\textbf{Overall Loss: }}\mathcal{L} &= \mathcal{L}_{CE} + \beta \mathcal{L}_{proto}  + \gamma \mathcal{L}_{cov} + \zeta \mathcal{L}_{CS} 
	\end{split}
\end{align}

\noindent where $\mathcal{L}_{CE}$ is the cross entropy loss, $\mathcal{L}_{proto}$ is the example-prototype loss in Eq. \ref{eq:4}, $\mathcal{L}_{cov}$ is the prototype weighted covariance loss, and $\mathcal{L}_{CS}$ is a global prototype cosine similarity loss.  The hyperparameters $\beta$, $\gamma$, and $\zeta$ are the relative weights of the appropriate loss components.  

\paragraph{Computation of $\mathcal{L}_{proto}$} \label{sec:proto}

In an online training setting where the prototypes, ``class-centers", of each category can be updated while training, we define the example-prototype loss function for an example $(\vx_n,y_n)$ as:

\begin{align} \label{eq:4}
	\mathcal{L}_{proto} = \sum_k^{K} \mathds{1}(k = y_n)\ ||\hat{g}(\vx_n) - \hat{\vp}_k||^2 
\end{align}

If we assume convergence of the above loss component \cite{bottouOpt}, we can derive two important results regarding a prototype and its category's examples.

\begin{lemma} \label{lem:2}
	
	For a sufficiently trained network with low empirical risk on $\mathcal{L}_{proto}$, the features of each resulting prototype feature vector, $\vp_{k}$ for class $k$ of dimension $J$, converge to $\vp_{k,j} = \frac{1}{N_k}\sum_{n=1}^{N_k} \ g(\vx_{n,k})_j$, where $N_k$ is the number of training examples from class $k$.
	
\end{lemma}

Lemma \ref{lem:2} states that if we optimize the loss component $\mathcal{L}_{proto}$, the resulting values of the prototype vectors in feature space will converge towards the class-mean feature vectors for each class.  Note that we compute the Euclidean distance between the normalized form of the vectors $g(\vx_n)$ and $\vp_k$, thus reducing the angular distance between intra-class features and their class mean prototypes.  See Section \ref{sec:AppLem2} for a full proof of Lemma \ref{lem:2}. 

Furthermore, optimizing $\mathcal{L}_{proto}$ equates to minimizing the squared residuals of each feature component between example features and class prototypes.  Since the prototypes converge to the class feature means, we can show an additional important aspect of prototype convergence:

\begin{corollary} \label{cor:2}
	If Lemma \ref{lem:2} holds, then minimizing $\mathcal{L}_{proto}$ is equivalent to minimizing the individual feature variances over the training examples, $\sum_{j=1}^{J} \mathrm{Var} (g(\vx_{n,k})_j)$ for each class $k$.
\end{corollary}

Corollary \ref{cor:2} formally states that if the prototypes represent the class mean feature vectors and we minimize the squared residuals between examples and prototypes, then we minimize the sample variance of each class' features.  This property is important because it minimizes the diagonal terms of the feature covariance matrix, $\text{tr}(\text{Cov}(\hat{v}_i \hat{p}_i, \hat{v}_j \hat{p}_j))$.  See Section \ref{sec:AppCor2} for a full proof of Corollary \ref{cor:2}. 

\paragraph{Computation of $\mathcal{L}_{CS}$} Per the derived inequality in Lemma \ref{lem:1}, we would need to maximize the global prototype dissimilarity in order to decrease the probability bound.  In practice, we prefer the training dynamics of minimizing the prototype similarities such that the loss has a lower bound of zero.  Further, the prototype similarity dependency is quadratic, implying that our loss should also have a quadratic form.  Given the current state of the prototype set $\left\{\vp_1, ... ,\vp_K\right\}$, we calculate $\mathcal{L}_{CS}$ as:

\begin{align}
	\begin{split} \label{eq:2}
		\mathcal{L}_{CS} &= \frac{1}{K(K-1)} \sum\limits_{i \ne j}^K (CS(\vp_i,\vp_j))^2\\
	\end{split}
\end{align}

In words, we compute the squared cosine similarity between all-pairs of class prototype vectors, excluding the diagonal, at every minibatch and attempt to minimize the mean.  By squaring the cosine similarity term, the gradients tend to focus on the most similar pairings of prototypes.

\paragraph{Computation of $\mathcal{L}_{cov}$} \label{sec:lcov} The most novel portion of the algorithm is the minimization of the CPR numerator (Defn. \ref{def:1})): $\mathbf{1}^T S_k \mathbf{1}$.  While $\mathcal{L}_{proto}$ minimizes the diagonal terms of $S_k$, the $\mathcal{L}_{cov}$ loss targets the off-diagonal terms of $S_k$. The computation of the covariance function for class $k$, $\mathrm{Cov}(\hat{v}_i \hat{p}_i, \hat{v}_j \hat{p}_j)_k = \E[(\hat{v}_i \hat{p}_i-\E[\hat{v}_i \hat{p}_i])(\hat{v}_j \hat{p}_j-\E[\hat{v}_j \hat{p}_j])]_k$, is simplified because, recall from theoretical framework, the prototypes are frozen for the computation of this loss component such that it acts as a vector of constants for $\mathcal{L}_{cov}$ only (prototypes still receive gradients from $\mathcal{L}_{proto}$ and $\mathcal{L}_{CS}$).  We drop the subscript $k$ for brevity, but emphasize that this loss component is computed only between an example's features and the prototype features of its true class.  The covariance function can be simplified as,
\begin{align}
	\begin{split} \label{eq:3}
		\mathrm{Cov}(\hat{v}_i \hat{p}_i, \hat{v}_j \hat{p}_j) &= \E[(\hat{v}_i \hat{p}_i-\hat{p}_i\E[\hat{v}_i])(\hat{v}_j \hat{p}_j-\hat{p}_j\E[\hat{v}_j )]\\
		&= \E[\hat{p}_i (\hat{v}_i - \E[\hat{v}_i])\hat{p}_j (\hat{v}_j -\E[\hat{v}_j])]\\
		&= \E[\hat{p}_i (\hat{v}_i - \hat{p}_i)\hat{p}_j (\hat{v}_j -\hat{p}_j)]\\
	\end{split}
\end{align}
Algorithm \ref{alg:lcov1} provides an efficient implementation to minimize the intra-class feature covariance terms in Eqn. \ref{eq:3}.  Utilizing Lemma \ref{lem:2}, we treat the learned class prototypes as the class-mean feature vectors and therefore as we optimize $\mathcal{L}_{proto}$, the prototypes constantly adjust to the mean of their class examples.  In mathematical terms, we can replace the term $\E[\hat{v_i}]$ with the current value of the appropriate prototype.

\begin{figure}[h]
	\centering
	\includegraphics[width=0.4\linewidth]{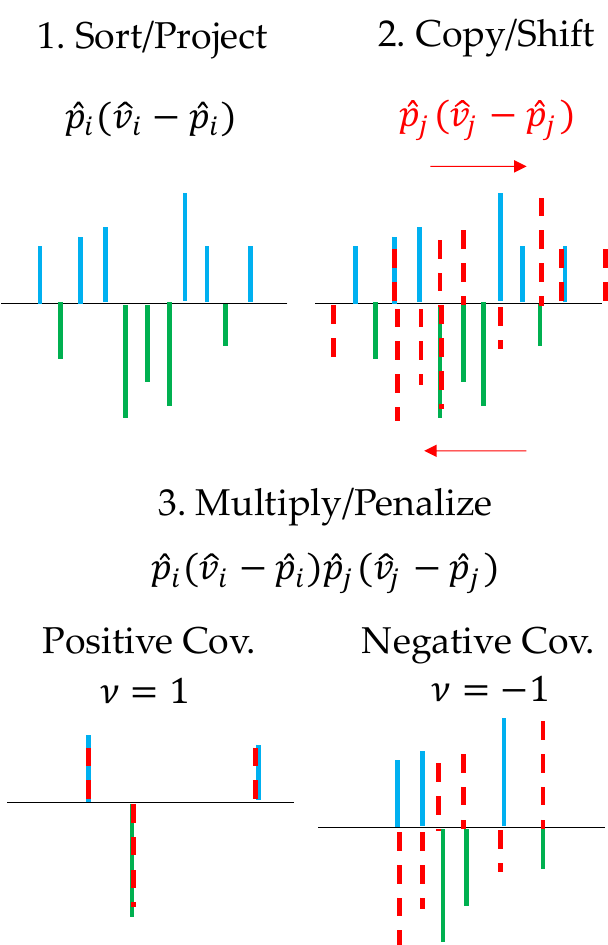}
	\caption[Lcov Algorithm]{Illustration of sorting and shifting strategy to optimize prototype weighted feature covariance in $\mathcal{O}(JlogJ)$ time.  Black line represents indices of class prototype sorted by its activation values; blue lines are example activations greater than prototype at corresponding index and green activations are smaller.  The value of $\nu$ can be selected to target positive (1), negative (-1), or both (0) signs of feature covariance.}
	\label{fig:covpic}
\end{figure}

Instead of calculating all possible off-diagonal terms of the $J\text{x}J$ feature covariance matrix in $\mathcal{O}(J^2)$ time, we compute an effective approximation in $\mathcal{O}(JlogJ)$ as illustrated in Fig. \ref{fig:covpic}.  Given an example, we first identify the correct prototype feature vector for the example based on its label, sort the features of the selected prototype (black line), reindex the examples' features by the sorted indices of their prototype, compute their activation differences at each index(blue and green lines) (step 1), randomly re-align (shift) these differences by padding with zeros (step 2), and then compute their padded element-wise product (step 3).  The user-parameter $\nu$ allows the loss term to regularize positive, negative, or both possible signs of off-diagonal covariance terms. The sorting operation is the most expensive in $\mathcal{O}(JlogJ)$ time and we use two linear (padded) copies of the features for $\mathcal{O}(J)$ space complexity.\linebreak

\begin{algorithm}	
	\caption{Computation of $\mathcal{L}_{cov,n}$ for single example, class $k$}\label{alg:1}
	\KwIn{$\vv_k = g(\vx_{n,k};\theta_{g}) \in \R^J$, $\vp_k \in \R^J$, $\nu$}
	\KwOut{$\mathcal{L}_{cov,n}$}
	$\hat{\vv_k} \gets normalize(\vv_k)$ \\
	$\hat{\vp_k} \gets normalize(\vp_k)$ \\
	$\hat{\vp_k} \gets sort(\hat{\vp_k})$\\
	$\hat{\vv_k} \gets reindex(\hat{\vv_k})$\\ \tcp{rearrange $\hat{\vv_k}$ by the sorted indices of $\hat{\vp_k}$}
	$\delta = \hat{\vp_k} \odot (\hat{\vv_k} - \hat{\vp_k})$  \tcp{same as $\hat{p}_i (\hat{v}_i - \hat{p}_i)$}
	$r \gets randint(1,10)$\quad \tcp{random integer from 1-10}
	
	$\delta_{pad,L} \gets PadLeftZeros(\delta,r)$\quad \tcp{Pad $r$ zeros on left}
	$\delta_{pad,R} \gets PadRightZeros(\delta,r)$\quad \tcp{Pad $r$ zeros on right}
	\uIf{$\nu == 0$}{
		$Z \gets |\delta_{pad,L} \odot \delta_{pad,R}|$\\
	}
	\Else{
		$Z \gets ReLU(sign(\nu)(\delta_{pad,L} \odot \delta_{pad,R}))$\\
	}
	
	return $\mathcal{L}_{cov,n} = \frac{1}{J+r}\sum_{j=1}^{J+r}Z$
	\label{alg:lcov1}
\end{algorithm}

\paragraph{Loss Summary} The total loss function in Eq. \ref{eq:5} works as a symbiotic system, each contributing to the following:

\begin{itemize}
	\item $\mathcal{L}_{CE}$ fits the classifier decision boundary to the training examples.
	\item $\mathcal{L}_{proto}$ maintains the class prototypes as the class mean feature vectors of their respective classes by Lemma \ref{lem:2} and thus makes the prototypes useful in the covariance calculations of $\mathcal{L}_{cov}$.  By Corollary \ref{cor:2}, this loss function also minimizes the diagonal terms of the intra-class covariance matrices.
	\item $\mathcal{L}_{cov}$ regularizes the off-diagonal terms of the prototype-weighted intra-class feature covariance matrices.
	\item $\mathcal{L}_{CS}$ reduces the global similarity between class prototypes.
	
\end{itemize}

\section{Empirical Results}

\subsection{Key Evaluation Method}

Previous efforts evaluate regularization techniques by training on the full available training set with different model initializations and report either a maximum or average test accuracy over model initializations.  We suggest that a robust way to assess overfitting performance is to randomly draw many different training subsets from the full available training set because we need to test whether the learning algorithm resists overfitting \emph{regardless of the seen examples}.  For our assessments, we randomly draw 12 training subsets from the available training data, ranging in size from 30\% to 50\% of the size of the source set.  We draw these training subsets in a stratified manner such that each category has the same number of samples.  For evaluating test performance, we test our selected model's accuracy on the 100\% of the provided test dataset.  We note that this evaluation method is mathematically consistent with the numerical definition of learning algorithm variance in bias-variance decomposition \cite{yangBiasVar}.

\subsection{Datasets, Architectures, Training}

We applied our algorithm and previous works to the well-known image classification data sets CIFAR100 \cite{CIFAR10} and STL10 \cite{STL10_Coates}.  We trained ResNet \cite{heResNet} based architectures for both datasets in a variety of widths to vary the number of available penultimate features, $J$.  We trained all models for 100 epochs using SGD (momentum=0.9) on a cosine annealed learning schedule beginning at a learning rate of 0.1, used a batch size of 128, and studied two different values of weight decay ($0.0$ and $5e-4$).  All runs were computed on a single GPU.  All model trainings included a 10 epoch warmup period before any regularizers were applied.  Standard flipping and cropping were employed for data augmentation in all runs. 

The historical regularization algorithms we chose for comparison were OrthoReg \cite{rodríguez2017regularizing}, DeCov \cite{cogswellDecov}, and Squentropy \cite{squent}.  For exCPR, we did not conduct a large hyperparameter study on ($\beta$,$\gamma$,$\zeta$, and $r$) due to the required computational resources, so we instead drew the 1st random training set and adjusted $\beta$,$\gamma$, and $\zeta$ such that our algorithm achieved 100\% training accuracy after 100 epochs while still minimizing the other loss components to the greatest possible extent.  The hyperparameter $r$ was set at 10 for each class and not adjusted afterward.  We then fixed the hyperparameters and used those for training and evaluation on the remaining 11 random training sets.

\subsection{Test Accuracy Results}

We ran a baseline (cross-entropy loss component only) and all regularization algorithms for two choices of weight decay (0.0, 5e-4) for various depths (ResNet18 and ResNet34) and widths ($J=64,96,256$) of the ResNet architecture \cite{heResNet}.  All results are reported as averages over 12 randomly drawn training sets (the same 12 are used for all algorithms).  We report test accuracy average, standard deviation, and minimum over all training sets.    

\begin{table}[h]
	\smaller
	\setlength\tabcolsep{4 pt} 
	\caption{CIFAR100, ResNet18 ($J=256$) Test Accuracy Averaged over 12 Random Training Sets of 50\% Nominal Size}
	\label{C100WD0main}
	\centering
	\rowcolors{4}{lightgray}{}
	\begin{tabular}{lcccccc}
		&\multicolumn{6}{c}{Test Accuracy} \\ \hline
		~ & Base & Decov & Ortho& Sqent & Ours & Ours \\ \hline
		~ & ~ & w=0.1 & ~ & ~ & $\nu$=0 & $\nu$= -1 \\ \hline
		&\multicolumn{6}{c}{Weight Decay = 0.0} \\ \hline
		$\mu$ & 0.594 & 0.607 & 0.595 & 0.616 & \B 0.622 & \B 0.622 \\ 
		$\sigma$ & 0.005 & 0.009 & 0.008 & 0.009 & 0.006 & 0.005 \\ 
		Min & 0.581 & 0.590 & 0.582 & 0.596 & 0.613 & \B 0.615 \\ \hline
		&\multicolumn{6}{c}{Weight Decay = 5e-4} \\ \hline
		$\mu$ & 0.661 & 0.667 & 0.662 & 0.661 &0.667 & \B 0.672 \\ 
		$\sigma$ & 0.002 & 0.003 & 0.004 & 0.004 &0.004 & 0.003 \\ 
		Min & 0.658 & 0.660 & 0.655 & 0.652 &0.660 & \B 0.668 \\ \hline
	\end{tabular}
\end{table}

\begin{table}[h]
	\smaller
	\setlength\tabcolsep{4 pt} 
	\caption{STL10, ResNet34 ($J=256$) Test Accuracy over 12 Random Training Sets of 50\% Nominal Size}
	\label{STL10WD0main}
	\centering
	\rowcolors{4}{lightgray}{}
	\begin{tabular}{lcccccc}
		&\multicolumn{6}{c}{Test Accuracy} \\ \hline
		~ & Base & Decov  & Ortho & Sqent & Ours & Ours \\ \hline
		~ & ~ & w=0.01  & ~ & ~ & $\nu$=0 & $\nu$= -1 \\ \hline
		&\multicolumn{6}{c}{Weight Decay = 0.0} \\ \hline
		$\mu$ & 0.632 & \B 0.638 & 0.613 & 0.576 & 0.634 & 0.633 \\ 
		$\sigma$ & 0.033 & 0.026 & 0.032 & 0.036 & 0.027 & 0.023 \\ 
		Min & 0.572 & 0.593 & 0.562 & 0.508 & 0.580 & \B 0.598 \\ \hline
		&\multicolumn{6}{c}{Weight Decay = 5e-4} \\ \hline
		$\mu$ & 0.640 & \B 0.664  & 0.642& 0.636 & 0.652 & 0.662 \\ 
		$\sigma$ & 0.040 & 0.019  & 0.036& 0.029 & 0.028 & 0.015 \\ 
		Min & 0.571 & 0.621  & 0.555& 0.577 & 0.583 & \B 0.632 \\ \hline
	\end{tabular}
\end{table}

Tables \ref{C100WD0main}, \ref{STL10WD0main}, \ref{CIFAR100M8}, and \ref{CIFAR100M12} show that our algorithm is effective in boosting generalization performance regardless of the training sets selected when compared to either the baseline or other regularization algorithms.  In a few settings, we have nearly identical performance to another regularization algorithm such as Decov or Squentropy, which we attribute to inoptimal hyperparameter settings for our algorithm and the fact that we elect to compute only a portion of the full covariance matrix to prioritize computation efficiency.  The results additionally demonstrate, as expected, that weight decay reduces overfitting across all algorithms regardless of setting.  Whether the regularization algorithms are more effective in a non-weight decay or weight decay setting is dependent on the dataset: for CIFAR100 the regularization algorithms are more effective without weight decay present, but the opposite is true for STL10.  For exCPR, we ran the algorithm targeting both positive/negative covariance terms ($\nu$=0) and also targeting only negative covariance terms ($\nu$=-1).  We discovered that targeting only negative covariances performed better in the weight decay setting and their performances were similar in non-weight decay.   Penalizing negative off-diagonal covariances is non-intuitive in the sense that this alone would actually \emph{increase} the CPR numerator.  However, we found that ($\nu$=-1) actually indirectly benefitted the diagonal terms of the feature covariance matrix and generalized this effect well to the test set (as can be found in Tables \ref{CIFAR100gap} and \ref{STL10gap}). 

\begin{table}[h]
	\smaller
	\setlength\tabcolsep{4 pt} 
	\caption{CIFAR100, ResNet34 ($J=64$) Test Accuracy over 12 Random Training Sets of 30\% Nominal Size}
	\label{CIFAR100M8}
	\centering
	\rowcolors{4}{lightgray}{}
	\begin{tabular}{lcccccc}
		\hline
		&\multicolumn{6}{c}{Test Accuracy} \\ \hline
		~ & Base & Decov & Ortho & Sqent & Ours & Ours \\ \hline
		~ & ~ & w=0.1 & ~ & ~ & $\nu$=0 & $\nu$=-1 \\ \hline
		&\multicolumn{6}{c}{Weight Decay = 0.0} \\ \hline
		$\mu$  & 0.430 & 0.421 & 0.423 & 0.440 & \B 0.445 & \B 0.445 \\ 
		$\sigma$ & 0.007 & 0.009 & 0.011 & 0.010 & 0.015 & 0.010 \\ 
		Min & 0.419 & 0.405 & 0.409 & 0.422 & 0.417 & \B 0.427 \\ \hline
		&\multicolumn{6}{c}{Weight Decay = 5e-4} \\ \hline
		$\mu$  & 0.517 & 0.515 & 0.514 & \B 0.520 & 0.515 & 0.518 \\ 
		$\sigma$ & 0.007 & 0.010 & 0.010 & 0.006 & 0.011 & 0.010 \\ 
		Min & 0.502 & 0.503 & 0.495 & \B 0.509 & 0.497 & 0.496 \\ \hline
	\end{tabular}
\end{table}

\begin{table}[h]
	\smaller
	\setlength\tabcolsep{4 pt} 
	\caption{CIFAR100, ResNet34 ($J=96$) Test Accuracy over 12 Random Training Sets of 30\% Nominal Size}
	\label{CIFAR100M12}
	\centering
	\rowcolors{4}{lightgray}{}
	\begin{tabular}{lcccccc}
		\hline
		&\multicolumn{6}{c}{Test Accuracy} \\ \hline
		~ & Base & Decov & Ortho & Sqent & Ours & Ours \\ \hline
		~ & ~ & w=0.1 & ~ & ~ & $\nu$=0 & $\nu$=-1 \\ \hline
		&\multicolumn{6}{c}{Weight Decay = 0.0} \\ \hline
		$\mu$ & 0.449 & 0.451 & 0.454 & 0.461 & \B 0.484 & 0.476 \\  
		$\sigma$ & 0.012 & 0.013 & 0.015 & 0.012 & 0.013 & 0.011 \\  
		Min & 0.433 & 0.435 & 0.426 & 0.447 & \B 0.460 & 0.451 \\  \hline
		&\multicolumn{6}{c}{Weight Decay = 5e-4} \\ \hline
		$\mu$ & 0.535 & 0.538 & 0.538 & 0.547 & 0.546 & \B 0.550 \\ 
		$\sigma$ & 0.009 & 0.009 & 0.008 & 0.007 & 0.007 & 0.006 \\ 
		Min & 0.519 & 0.519 & 0.523 & \B 0.536 & 0.535 & 0.535 \\ \hline
	\end{tabular}
\end{table}

\subsection{CPR Generalization Results}

We disassemble the CPR metric (Defn. \ref{def:1}) and measure how well the intra-class covariance and inter-class separation terms ($DS^2$) optimize on the training set and generalize to the test set for all algorithms.  Tables \ref{CIFAR100gap} and \ref{STL10gap} show that exCPR reduces prototype-weighted intra-class covariance on the training set significantly compared to other algorithms.  The training set feature covariance generalizes to the test set to varying degrees.  The $\nu$=-1 implementation of exCPR generalizes feature variation significantly better to the test set than $\nu$=0, but the opposite is true in regards to inter-class prototype separation ($DS^2$).  This implies that the best, most generalizable setting of $\nu$ may differ depending on the class and feature in question.

\begin{table}[h]
	\smaller
	\setlength\tabcolsep{3 pt} 
	\setlength\extrarowheight{0pt}
	\caption{CIFAR100, ResNet18 ($J=256$) CPR Components Averaged over 12 Random Training Sets of 50\% Nominal Size}
	\label{CIFAR100gap}
	\centering
	\rowcolors{4}{}{lightgray}
	\begin{tabular}{lccccccc}
		\hline
		&\multicolumn{3}{c}{$\frac{1}{K}\sum\limits^{K} \mathbf{1}^T S_k \mathbf{1}$}  & \multicolumn{3}{c}{$DS^2$}  & ~ \\ \hline
		~ & Train & Test & Gap & Train & Test & Gap & Acc \\ \hline
		&\multicolumn{6}{c}{Weight Decay = 0.0} \\ \hline
		Base & 0.0030 & \B 0.0041 &  0.0011 & 0.132 & 0.089 & 0.043 & 0.594 \\ 
		Ortho & 0.0030 & 0.0045 & 0.0015 & 0.132 & 0.097 & \B 0.036 & 0.595 \\ 
		Decov & 0.0024 & 0.0054 & 0.0030 & 0.226 & 0.105 & 0.121 & 0.600 \\ 
		Sqent & 0.0027 & 0.0986 & 0.0959 & \B 0.973 & \B 0.909 & 0.064 & 0.616 \\ 
		$\nu$=0& \B 0.0015 & 0.0286 & 0.0271 & 0.720 & 0.571 & 0.149 & \B 0.622 \\ 
		$\nu$=-1& 0.0017 & 0.0053 & 0.0036 & 0.255 & 0.190 & 0.065 & \B 0.622 \\ \hline
	\end{tabular}
	\begin{tabular}{lccccccc}
		&\multicolumn{6}{c}{Weight Decay = 5e-4} \\ \hline
		Base & 0.0016 & \B 0.0056 &  0.0041 & 0.251 & 0.156 & 0.095 & 0.661 \\ 
		Ortho & 0.0016 & 0.0062 & 0.0046 & 0.252 & 0.178 & \B 0.074 & 0.662 \\ 
		Decov & 0.0015 & 0.0151 & 0.0136 & 0.547 & 0.287 & 0.260 & 0.667 \\ 
		Sqent & 0.0012 & 0.0636 & 0.0624 & \B 0.917 & \B 0.727 & 0.190 & 0.661 \\ 
		$\nu$=0 & \B 0.0005 & 0.0319 & 0.0314 & 0.764 & 0.598 & 0.166 & 0.667 \\ 
		$\nu$=-1& 0.0008 & 0.0102 & 0.0094 & 0.453 & 0.305 & 0.148 & \B 0.672 \\ \hline
	\end{tabular}
\end{table}

\begin{table}[h]
	\smaller
	\setlength\tabcolsep{3 pt} 
	\setlength\extrarowheight{0pt}
	\caption{STL10, ResNet34 ($J=256$) CPR Components over 12 Random Training Sets of 50\% Nominal Size}
	\label{STL10gap}
	\centering
	\rowcolors{4}{}{lightgray}
	\begin{tabular}{lccccccc}
		\hline
		&\multicolumn{3}{c}{$\frac{1}{K}\sum\limits^{K} \mathbf{1}^T S_k \mathbf{1}$}  & \multicolumn{3}{c}{$DS^2$}  & ~ \\ \hline
		~ & Train & Test & Gap & Train & Test & Gap & Acc \\ \hline
		&\multicolumn{6}{c}{Weight Decay = 0.0} \\ \hline
		Base & 0.0010 & 0.0024 & 0.0014 & 0.069 & 0.038 & 0.031 & 0.632 \\ 
		Ortho  & 0.0010 & 0.0023 & 0.0013 & 0.065 & 0.036 & 0.030 & 0.613 \\ 
		Decov  & 0.0004 & \B 0.0010 & 0.0006 & 0.024 & 0.007 & 0.017 & \B 0.638 \\ 
		Sqent  & 0.0090 & 0.0207 & 0.0117 & \B 0.417 & \B 0.214 & 0.203 & 0.576 \\ 
		$\nu$=0  & 0.0003 & 0.0045 & 0.0042 & 0.227 & 0.179 & 0.048 & 0.634 \\ 
		$\nu$=-1  & \B 0.0001 & 0.0011 & 0.0010 & 0.067 & 0.048 & 0.018 & 0.633 \\ \hline
	\end{tabular}
	\begin{tabular}{lccccccc}
		&\multicolumn{6}{c}{Weight Decay = 5e-4} \\ \hline
		Base  & 0.0009 & 0.0032 & 0.0023 & 0.114 & 0.061 & 0.054 & 0.640 \\ 
		Ortho  & 0.0009 & 0.0032 & 0.0022 & 0.115 & 0.061 & 0.054 & 0.642 \\ 
		Decov  & 0.0003 & \B 0.0014 & 0.0011 & 0.042 & 0.011 & 0.031 & \B 0.664 \\ 
		Sqent  & 0.0094 & 0.0242 & 0.0148 & \B 0.497 & \B 0.287 & 0.209 & 0.633 \\ 
		$\nu$=0  & 0.0002 & 0.0046 & 0.0045 & 0.254 & 0.188 & 0.066 & 0.652 \\ 
		$\nu$=-1  & \B 0.0001 & 0.0015 & 0.0014 & 0.098 & 0.068 & 0.030 & 0.662 \\ \hline
	\end{tabular}
\end{table}

\paragraph{Discussion} Our CPR metric provides the relative numerical importance between intra-class feature covariance (weighted by prototype activations) and inter-class separation (quadratic in prototype dissimilarity).  While we wanted to show that explicitly reducing CPR in our algorithm improves overfitting over an unregularized baseline, we critically wanted to show that other non-CPR regularization techniques still implicitly improved CPR and resulted in better test performance, which would confirm our probabilistic model’s relationship to misclassification (reducing the CPR for each class reduces misclassification risk).  We chose OrthoReg, DeCov, and Squentropy because they are simple, effective, and address variation and separation differently by either the weights (OrthoReg), feature activations (DeCov), or logits (Squentropy). 

Our probabilistic model for CPR has parallels to neural collapse property (NC1), which is the collapse of within-class feature covariance when a neural network has been trained for a sufficient number of epochs beyond 100\% training accuracy (see \cite{Papyan_2020} for details).  Through this lens, our work shares a connection to neural collapse in that our algorithm significantly reduces intra-class feature covariance compared to baseline models within a typical duration (100 epochs) of training, which is not in the terminal phase of training – our loss components thus accelerate the covariance effects of neural collapse, which has been shown to improve generalization in many settings and lends credence to our Chebyshev model that reducing CPR also reduces overfitting \cite{Papyan_2020}. 

We highlight some important differences and advantages of our method. First, our probability model informs us on the specific mathematical forms of intra-class covariance vs. class separation. Our feature covariance is uniquely prototype-weighted since we use $\E[p_i(v_i-p_i)p_j (v_j-p_j)]$.  All categories have differently shaped prototype feature vectors and thus our weighted covariance will scale each feature gradient differently in backpropagation depending on its relative importance to each class.  We further wanted to design a covariance algorithm that reduces the computational complexity.  Using our feature sorting and padding algorithm, our method computes the covariance contributions for a sample in $\mathcal{O}(JlogJ)$ time, in comparison to previous approaches in $\mathcal{O}(J^2)$ time \cite{cogswellDecov}. Our results show that over the course of training, this method still reduces model feature covariance even though we do not explicitly compute the full covariance matrix.  This gives our algorithm a distinct scaling advantage.

\subsection{Conclusion}
\label{sec:conclusion}

Many efforts for overfitting reduction to improve test performance have been shown effective.  It has been shown that overfitting can be reduced by amending the standard cross-entropy loss with terms in one or more hidden layers of the network, including the convolutional layers. To build upon the wisdom guiding these previous works, we analytically tried to understand how class feature relationships affect misclassification.
\par 
Our analysis began by assuming a DNN is the composition of a feature extractor and classifier, where the classifier is the last fully connected layer of the network and the feature layer is the input vector to the classifier. Assuming that, corresponding to each class, there exists an ideal feature vector which we designate as a class prototype. 

\par
Formally, we derived Chebyshev and Chebyshev-Cantelli probability bounds on the deviation of cosine similarity between the features of an example and its class prototype.
We added the terms in the inequality to define our novel loss function for optimizing the feature layer, but the new loss function backpropagates errors to the previous convolutional layers' and optimizes their parameter values as well. The new loss function based on our probability bounds effectively reduces intra-class feature covariance while keeping class examples separated in feature space, which reduces the risk of overfitting.
Empirical results on multiple datasets and network architectures validates that the our loss function reduces overfitting and improves upon previous approaches in an efficient manner.

\bibliographystyle{plain}
\bibliography{citations}

\section{Appendix}

\subsection{Proof of Lemma \ref{lem:1}}
\label{sec:AppLem1}
\begin{proof}
	For a non-negative random variable $X$, Markov's inequality states:
	\begin{align}
		\begin{split} \label{eq:35}
			Pr[X \ge a] &\le \frac{\E[X]}{a}\\
		\end{split}
	\end{align}

	Let the non-negative random variable $X$ be $(Y - E[Y])^2 \ge 0$ where $Y \in \R$ is some real-valued random scalar variable and let $a = c^2$ where c is some positive real-valued scalar, then Markov's inequality becomes,

	\begin{align}
		\begin{split} \label{eq:36}
			Pr[(Y - E[Y])^2 \ge c^2] &\le \frac{\E[(Y - E[Y])^2]}{c^2}\\
			Pr[(Y - E[Y])^2 \ge c^2] &\le \frac{\text{Var}(Y)}{c^2}\\
			Pr[|Y - E[Y]| \ge c] &\le \frac{\text{Var}(Y)}{c^2}\\
		\end{split}
	\end{align}
	
	The last inequality in Eqn. \ref{eq:36} is the two-tailed form of Chebyshev's inequality.  We now choose a specific scalar random variable for $Y$ that is applicable to our problem of interest.  Given a prototype feature vector $\vp$ for class $k$ and the feature vector of an example from class $k$, $\vv_k = g(\vx_k)$, we consider the scalar random variable $Y$ to be the cosine similarity between the example's feature vector and it's class prototype:
	\begin{align}
		\begin{split} \label{eq:37}
			Y  = \frac{\vv_k \cdot \vp_k}{||\vv_k||||\vp_k||} \defeq CS(\vv_k,\vp_k)
		\end{split}
	\end{align}
	
	We assume that the cosine similarity is a summation of correlated random variables with each variable being the product of a unit vector component of  $\vp_k$ and $\vv_k$ at feature $j$,
	\begin{align} \label{eq:38}
		Y = \frac{\sum_j^J v_{k,j} p_{k,j}}{||\vv_k||||\vp_k||}
	\end{align}
	
	Furthermore, we can think of the summation as the sampling of two unit vectors $\hat{\vv}_k$ and $\hat{\vp}_k$ from the unit hypersphere so that X becomes the summation of unit vector components, each component being a random variable:
	\begin{align} \label{eq:39}
		Y = \sum_j^J \hat{v}_{k,j} \hat{p}_{k,j}
	\end{align}
	
	Before substituting into Eq. \ref{eq:36}, we expand the expression $\text{Var}(Y)$.  The general form of Bienaymé's identity states \cite{papoulisProb}:
	\begin{equation} \label{eq:39a}
		\begin{split}
			\text{Var}(Y) =\text{Var}(CS(\vv_k,\vp_k)) = \text{Var}(\sum_j^J \hat{v}_{k,j} \hat{p}_{k,j}) = \sum\limits_{i,q = 1} \text{Cov}(\hat{v}_{k,i} \hat{p}_{k,i},\hat{v}_{k,q} \hat{p}_{k,q})
		\end{split}
	\end{equation}
	
	This identity introduces the \emph{covariance} function as a generalized form of variance between two random variables.  In our case, we can define the elements of a symmetric positive semi-definite covariance matrix $S_k \in \R^{J \times J}$,
	\begin{equation} \label{eq:39b}
		\begin{split}
			S_{k,i,q} = \text{Cov}(\hat{v}_{k,i} \hat{p}_{k,i},\hat{v}_{k,q} \hat{p}_{k,q}) = \E_x [ (\hat{v}_{k,i} \hat{p}_{k,i} - \E_x [\hat{v}_{k,i} \hat{p}_{k,i}]) (\hat{v}_{k,q} \hat{p}_{k,q} - \E_x[\hat{v}_{k,q} \hat{p}_{k,q}]) ]
		\end{split}
	\end{equation}
	
	\noindent where $\E_x$ indicates the expectation is taken over the data samples.  Because $S_k$ is positive semi-definite, it satisfies ${\vu}^T S_k \vu \ge 0$ for all real vectors $\vu \in \R^J$.  Utilizing $S_k$, we can convert Eq. \ref{eq:39a} to matrix form,
	\begin{align} \label{eq:39c}
		\sum\limits_{i,q = 1} \text{Cov}(\hat{v}_{k,i} \hat{p}_{k,i},\hat{v}_{k,q} \hat{p}_{k,q}) = \mathbf{1}^T \text{Cov}(\hat{v}_{k,i} \hat{p}_{k,i},\hat{v}_{k,q} \hat{p}_{k,q}) \mathbf{1} = \mathbf{1}^T S_k \mathbf{1}
	\end{align}
	
	We substitute our derived quantities into Eq. \ref{eq:36} in separate steps for clarity.  First we replace the scalar random variable $Y$ by $CS(\vv_k,\vp_k)$ as defined in Eq. \ref{eq:37}:
	\begin{align}
		\begin{split} \label{eq:39d}
			Pr[|CS(\vv_k,\vp_k) - E[CS(\vv_k,\vp_k)]| \ge c] &\le \frac{\text{Var}(CS(\vv_k,\vp_k))}{c^2}\\
		\end{split}
	\end{align}
	
	Next, we substitute in our matrix form expression for $\text{Var}(CS(\vv_k,\vp_k))$ from Eq. \ref{eq:39c}:
	\begin{align}
		\begin{split} \label{eq:39e}
			Pr[|CS(\vv_k,\vp_k) - E[CS(\vv_k,\vp_k)]| \ge c] &\le \frac{\mathbf{1}^T S_k \mathbf{1}}{c^2}\\
		\end{split}
	\end{align}
	
	Lastly, our only numerical requirement on $c$ is that it is a real valued, positive scalar.  We define the mean dissimilarity between class $k$ and all other $K-1$ class prototypes as follows and select it as a meaningful value for $c$ to represent the average angular distance between class $k$ and all other classes:
	
	\begin{align}
		\begin{split} \label{eq:41}
			DS_k &\defeq 1 - \frac{1}{(K-1)} \sum\limits_{i \ne k}^{K-1} CS(\vp_k,\vp_i)\\
		\end{split}
	\end{align} 
	
	We therefore select $DS_k \in (0,1] $ as a meaningful value for $c$, where we exclude the case where $DS_k = 0$, which is equivalent to all class prototype feature vectors being parallel.  By assuming that the network is trained sufficiently, this case is excluded.  Substituting $DS_k$ for $c$ we reach our final two-tailed Chebyshev inequality:
	\begin{align}
		\begin{split} \label{eq:40}
			Pr\big[\big|CS(\vv_k,\vp_k) - &\E\big[CS(\vv_k,\vp_k)\big]\big| \ge DS_k\big] \le \frac{\mathbf{1}^T S_k \mathbf{1}}{DS_k^2}\\
		\end{split}
	\end{align}

	Thus concludes our proof.
\end{proof}

\subsection{Proof of Corollary \ref{cor:1}}
\label{sec:AppCor1}
\begin{proof}
	The proof of the one-sided version of our inequality follows the procedure detailed by \cite{boucheronInequality}.
	
	Define a new random variable $Z$:
	\begin{align}
		\begin{split} \label{eq:13b}
			Z &= Y - \E[Y] \\
		\end{split}
	\end{align} 
	
	The mean and variance of this random variable can be directly deduced:
	\begin{align}
		\begin{split} \label{eq:13c}
			\E[Z] &= \E[Y - \E[Y]] = \E[Y] - \E[Y] = 0 \\
			\text{Var}(Z) &= \E[(Z - \E[Z])^2] = \E[(Z)^2] = \E[(Y - \E[Y])^2] = \text{Var}(Y)\\
		\end{split}
	\end{align} 
	Take the real-valued random variable $Y$ as before in Lemma \ref{lem:1}:
	\begin{align}
		\begin{split} \label{eq:13d}
			Y &= CS(\vv_k, \vp_k)
		\end{split}
	\end{align}
	Therefore, $\text{Var}(Z)$ can be re-expressed in terms of $CS(\vv_k, \vp_k)$:
	\begin{align}
		\begin{split} \label{eq:13e}
			\text{Var}(Z) &= \text{Var}(Y) = \mathbf{1}^T S_k \mathbf{1}
		\end{split}
	\end{align}
	where we have made use of Bienaymé's identity \cite{papoulisProb} and Eqn. \ref{eq:39a}.
	
	Let $u$ be a non-negative real valued scalar and bound the probability that $Z$ is greater than some value $\lambda$:
	\begin{align}
		\begin{split} \label{eq:13f}
			Pr(Y - \E[Y] \ge \lambda) &= Pr(Z \ge \lambda) \\
			&= Pr(Z + u \ge \lambda + u) \\
			&\le Pr((Z + u)^2 \ge (\lambda + u)^2) \\
			&\le \frac{\E[(Y + u)^2]}{(\lambda + u)^2} 	\\
			&\le \frac{\mathbf{1}^T S_k \mathbf{1} + u^2}{(\lambda + u)^2}				
		\end{split}
	\end{align}
	where the last two inequalities are due to Markov's inequality.  We note that there are no limitations on $u$ as long as it is non-negative and thus we can minimize the bound using an optimal value of $u$.  The optimal value of $u$ to minimize the bound is $u = \mathbf{1}^T S_k \mathbf{1} / \lambda$.  Therefore, the final form of the one-sided bound becomes:
	\begin{align}
		\begin{split} \label{eq:13g}
			Pr(Y - \E[Y] \ge \lambda) &\le \frac{\mathbf{1}^T S_k \mathbf{1}}{\mathbf{1}^T S_k \mathbf{1} + \lambda^2}				
		\end{split}
	\end{align}
	For our problem, we substitute in our cosine similarity measure for $Y$ and $DS_k$ for $\lambda$.  We are also more interested in bounding the lower tail where typically $Y < \E[Y]$, so we adjust change the direction of the inequality and arrive at our final single-tailed Chebyshev-Cantelli inequality:
	\begin{align}
		\begin{split} \label{eq:13h}
			Pr(Y - \E[Y] \le -DS_k) &\le \frac{\mathbf{1}^T S_k \mathbf{1}}{\mathbf{1}^T S_k \mathbf{1} + DS_k^2}				
		\end{split}
	\end{align}
	This concludes our proof.
\end{proof}

\subsection{Proof of Lemma \ref{lem:2}}
\label{sec:AppLem2}
\begin{proof}

	We begin our proof by recalling the work of Bottou et al. \cite{bottouOpt} on the convergence of general scalar loss objectives.  Specifically, we state a corollary from Bottou et al. that shows, under the stochastic gradient descent learning rule with diminishing step sizes, the expectation of the squared magnitude of the gradient vector of the loss converges to zero as the number of gradient steps approaches infinity (and the step sizes diminishes towards zero).  
	
	\begin{corollary} \label{cor:3}
		[From Bottou et al. \cite{bottouOpt}] Assuming a general scalar objective loss function $\mathcal{L}(\theta_z)$ that is twice differentiable and has Lipschitz-continuous derivatives under the mapping $\theta \mapsto \norm{ \nabla \mathcal{L}(\theta_z)}$ where $\theta_z$ is a learnable set of parameters at iteration $z$ of a stochastic gradient descent rule with diminishing step sizes, then,
		\begin{align}
			\begin{split}
				\lim\limits_{z \rightarrow \infty}\E\big[\norm{ \nabla \mathcal{L}(\theta_z)}^2 \big] = 0
			\end{split}
		\end{align}
	\end{corollary}
	
	The corresponding stochastic gradient descent update rule that applies to Corollary \ref{cor:3} is,
	
	\begin{align}
		\textrm{\textbf{Stochastic Gradient Descent Update: }} & \theta_{z+1} \leftarrow \theta_{z} - \alpha_z \nabla_{\theta} \mathcal{L}(\theta_z) \label{gradientUpdate}\\
	\end{align} 
	
	Because the Corollary \ref{cor:3} holds for any general objective function being optimized under stochastic gradient with diminishing stepsizes, we can apply it to our prototype objective function, which we restate here for a single example $(\vx_n,y_n)$ and $\vp_k \in \R^J$:
	
	\begin{align} \label{eq:96}
		\mathcal{L}_{proto} = \sum_k^{K} \mathds{1}(k = y_n)\ \norm{g(\vx_n) - \vp_k)}^2 
	\end{align}

	By Corollary \ref{cor:3}, we assume $\mathcal{L}_{proto}$ converges such that:
	
	\begin{align} \label{eq:98}
		\norm{\frac{1}{N}\sum_{n=1}^{N} \nabla \mathcal{L}_{proto} } = 0  
	\end{align}
	
	We now rewrite the expression on the left hand side.  We can expand Eqn. \ref{eq:98} using Eqn. \ref{eq:96} and then re-write $\norm{g(\vx_n) - \vp_k)}^2$ as a summation over its $J$ elements:
	
	\begin{align} \label{eq:99}
		\begin{split}
			\norm{\nabla \frac{1}{N}\sum_{n=1}^{N} \sum_{j=1}^{J} \mathds{1}(k = y_n)\ (g(\vx_n)_j - p_{k,j})^2 }  \\
		\end{split}
	\end{align}

	The outer norm, $||\cdot||$, can represent a summation over the terms of the gradient vector.  We replace the indicator function by subscripting $\vx$ with $k$ to indicate it has true label $k$. 	
	\begin{align} \label{eq:99b}
		\begin{split}
			\norm{\nabla \frac{1}{N_k}\sum_{n=1}^{N_k} \sum_{j=1}^{J}  (g(\vx_{n,k})_j - p_{k,j})^2 }  \\
		\end{split}
	\end{align}
	
	where $N_k$ is the number of examples in class $k$.
	
	At convergence, let us fix some feature extraction function $g(\theta_g)$ and compute the $J$ gradient vector terms with respect to the learnable elements of class prototype $\vp_k$.  The $j^{th}$ element of the gradient vector is:
	\begin{align}  \label{eq:99c}
		\begin{split}
			\norm{\frac{\partial}{\partial{p_{k,j}}}\frac{1}{N_k}\sum_{n=1}^{N_k} \sum_{j=1}^{J} (g(\vx_{n,k})_j - p_{k,j})^2}
		\end{split}
	\end{align}

	We now expand the outer norm as a summation, reintroduce the equality to zero, and square both sides:
	\begin{align}  \label{eq:100}
		\begin{split}
			\sum_{j=1}^{J}\left(\frac{\partial}{\partial{p_{k,j}}}\frac{1}{N_k}\sum_{n=1}^{N_k} \sum_{j=1}^{J} (g(\vx_{n,k})_j - p_{k,j})^2\right)^2 = 0
		\end{split}
	\end{align}
	
	Applying each partial derivative, we get,
	\begin{align}  \label{eq:101}
		\begin{split}
			\sum_{j=1}^{J}\left(\frac{1}{N_k}\sum_{n=1}^{N_k} \sum_{j=1}^{J} 2(p_{k,j} - g(\vx_{n,k})_j)\right)^2  = 0
		\end{split}
	\end{align}
	
	We observe that equations $\frac{1}{N_k}\sum_{n=1}^{N_k} 2(p_{k,j} - g(\vx_{n,k})_j) = 0,\quad j=1...J$ are a solution to Eqn. \ref{eq:101} and then find an expression for $g(\vx_{n,k})_j$:
	
	\begin{align} \label{eq:102}
		\begin{split}
			-p_{k,j} + &\frac{1}{N_k}\sum_{i=1}^{N_k} g(\vx_{n,k})_j = 0\ \ \ j = 1...J \\
			p_{k,j} &= \frac{1}{N_k}\sum_{i=1}^{N_k} g(\vx_{n,k})_j\ \ \ j = 1...J \\
		\end{split}
	\end{align}
	
	This concludes our proof.
\end{proof}

\subsection{Proof of Corollary \ref{cor:2}}
\label{sec:AppCor2}
\begin{proof}
	We assume that we apply minimize the empirical risk of $\mathcal{L}_{proto}$ such that Eqn. \ref{eq:102} applies and then we restate $\mathcal{L}_{proto}$:
	\begin{align} \label{eq:103}
		\begin{split}
			\frac{1}{N_k}\sum_{n=1}^{N_k} \mathcal{L}_{proto} = \frac{1}{N_k}\sum_{n=1}^{N_k} ||g(\vx_{n,k}) - \vp_k||^2  \\
		\end{split}
	\end{align}
	
	We write $||g(\vx_{n,k}) - \vp_k||^2$ as a summation over its $J$ elements and substitute in Eqn. \ref{eq:102} for $\vp_k$ to find a concise expression for the empirical risk of $\mathcal{L}_{proto}$.
	
	\begin{align} \label{eq:105}
		\begin{split}
			\frac{1}{N_k}\sum_{n=1}^{N_k} \mathcal{L}_{proto}&= \sum_{j=1}^{J}\frac{1}{N_k}\sum_{n=1}^{N_k} \left(g(\vx_{n,k})_j - p_{k,j}\right)^2  \\
			&= \sum_{j=1}^{J}\frac{1}{N_k}\sum_{n=1}^{N_k} \left(g(\vx_{n,k})_j - \frac{1}{N_k}\sum_{n=1}^{N_k} g(\vx_{n,k})_j\right)^2  \\
			&= \sum_{j=1}^{J} \mathrm{Var}(g(\vx_{n,k})_j)\ \   k=1...K \\
		\end{split}
	\end{align}
	
	where $\mathrm{Var}(g(\vx_{n,k})_j)$ are the individual feature vector sample variances over the training examples.
	
\end{proof}  

\end{document}

%% file: math_commands.tex
\usepackage{amsmath,amsfonts,bm}

\def\eqref#1{equation~\ref{#1}}

\def\1{\bm{1}}

\def\vp{{\bm{p}}}

\def\vu{{\bm{u}}}
\def\vv{{\bm{v}}}

\def\vx{{\bm{x}}}
\def\vy{{\bm{y}}}
\def\vz{{\bm{z}}}

\DeclareMathAlphabet{\mathsfit}{\encodingdefault}{\sfdefault}{m}{sl}
\SetMathAlphabet{\mathsfit}{bold}{\encodingdefault}{\sfdefault}{bx}{n}

\newcommand{\E}{\mathbb{E}}

\newcommand{\R}{\mathbb{R}}

